# Right Choice of Classification Algorithms Based on Reinforcement Learning for Prediction of Non-Alcoholic Fatty Liver

Hasan Samadbin[1], Arman Daliri[1]

Department of Computer Engineering, Karaj Branch, Islamic Azad University, Karaj, Iran

Hasan.Samadbin@Kiau.ac.ir

Arman.Daliri@kiau.ac.ir

***Abstract***

**There are many complex issues in the world of artificial intelligence. Some of these problems are solved using other artificial intelligence methods, which are called artificial intelligence for artificial intelligence. Finding an appropriate classifier algorithm is a time-consuming task. For this reason, an algorithm that can automatically learn the choice of classification algorithms is very important. Classification algorithms are useful in predicting various diseases. Also, Primary Biliary Cirrhosis is one of the most well-known diseases that have been predicted by classification algorithms. This research's most significant achievement and novelty is the automatic increase in learning through a scoring method of reinforcement learning is called square learning (SL). In this research, an algorithm is presented that learns to automatically select the appropriate classification algorithm to predict Primary Biliary Cirrhosis. In this article, with inspiration from four evaluation metrics in classification algorithms, a new reinforcement learning method by the name of Fourth Degree Learning has been presented. In this research, we increased the performance of the classification algorithms used in this method from 63% of accuracy and achieved 98% accuracy.**

## 1. INTRODUCTION

People are struggling with fatty liver disease (FLD) and its harmful effects these days. Liver diseases cause approximately two million human deaths worldwide every year [1]. Early prevention of Fatty liver using machine learning algorithms is common. Fatty liver is one of the most important causes of this condition, including alcoholic fatty liver and non-alcoholic Fatty liver [2]. One of the non-alcoholic chronic liver diseases is a primary biliary liver disease [3]. This disease is chronic, which means that it affects people's lives over time until the person eventually gets liver cancer [4]. For this reason, the prevention of this disease is essential.

A medical solution to treat Primary Biliary Cirrhosis (PBC) is monitored at home or hospital under the supervision of a doctor. There is no specific treatment for this disease, and doctors manage this chronic disease by prescribing medicine [5]. One of the practical solutions is the use of Ursodiol, which helps to eliminate bile in the liver [6]. All medical solutions are also dependent on the checkup test for the liver. Therefore, there is a serious need for machine learning methods to diagnose patients in this field. Machine learning algorithms have the ability to predict disease, but due to the existence of different datasets, the prediction accuracy may decrease [7].

Each machine learning algorithm performs differently from the others. For prediction, researchers deal with these issues by using supervised algorithms [8]. There are many classifier algorithms for PBC prediction and it is not possible to use all of them one by one [9]. Every day, researchers present new classifiers that are more likely to perform better than old versions. [10]. The applications of these algorithms are also in various issues such as multimedia [12] ,[11], improving selection in learning algorithms [13], detecting additional data in the network [14], multi-agent gaming problems [15], and medicine [16]. If this is the assumption that the required data is suitable for the algorithms and the algorithms work flawlessly, there is still a problem in selecting appropriate algorithms.
In this research, an algorithm that can learn to choose the classifier has been presented. This learning is to choose the appropriate classification algorithm to diagnose PBC Based on the mayo clinic PBC dataset [17] and three well-known classification algorithms which are Decision Tree [18], Random Forest Classifier [19] and Support Vector Classifier [20]. A new framework is implemented that is able to understand which algorithm is more suitable than the other. The source of inspiration for this framework is reinforcement learning methods. To the best of our knowledge, no such study incorporating reinforcement learning for choosing the appropriate classification algorithm to diagnose PBC has been carried out. The contribution of this work is twofold. First, we introduce a reinforcement learning approach, denoted as Fourth Degree Learning which addresses the problem of classification selection. Second, we utilized the proposed approach for automatic classification and prediction of PBC.
In the continuation of this research, the second section examines primary hepatobiliary disease, stating its importance. After that, the third section presents the new solution, and its different parts are explained. Then, in the fourth section, the test and evaluation results of the new method are discussed. Finally, the fifth section is the conclusion and the future works in this field.

## 2. RESEARCH BACKGROUND

*Primary biliary cirrhosis* is a disease that damages the liver [21]. This disease is chronic, meaning it remains for a long time or recurs regularly [4]. The bile ducts of people with primary biliary cholangitis are damaged [22]. Bile ducts are tiny tubes in the liver that transport bile from the liver to other body parts [23]. This accumulation of bile in the liver causes ulcers called cirrhosis [24]. Primary biliary liver disease is a progressive disease, which means that the patient's condition worsens over time. Left untreated can lead to liver cirrhosis, liver failure, and even death [25].
Many people with PBC in the early stages have no symptoms [26]. Some people find out they have it when their doctor checks them for another problem [26]. The general symptoms of this disease can be several examples, which include: fatigue, skin itching, abdominal pain, dark skin, small white or yellow bumps under the skin around the eyes, dry eyes and mouth, and muscle and joint pain [27]. Any other disease may cause the said symptoms; with the progress of this disease in the liver, more specific symptoms appear. The symptoms of PBC in the later stages are yellow skin and white eyes swelling of the feet, knees and legs, swelling of the abdomen due to fluid accumulation, internal bleeding in the upper part of the stomach or lower esophagus caused by giant vessels, nausea, weight loss and dark urine [28].
To propose such predictive systems, the first requirement is information and data [29]. For this reason, a reliable dataset in the field of PBC disease is used in this article. For this reason, an original dataset in the field of PBC disease is used in this article, which is called the Mayo Clinic trial in primary biliary cirrhosis of the liver [30]. This dataset includes 424 patients who visited the Mayo Clinic in 10 years between 1974 and 1984 [17]. One of the critical points of this dataset is the registration of deaths after the addition of patients, which can be used for time series analysis [30]. The Mayo dataset consists of twenty medical and clinical features, which are very suitable for applying machine learning methods. A more detailed analysis of this research has been done in the following.

## 3. METHODOLOGY OF RIGHT CHOICE OF CLASSIFICATION ALGORITHMS

In this research, a reinforcement learning algorithm is presented for the early prediction of PBC disease. This method is inspired by a voting system and consists of three main parts. In the first part of this framework, the dataset is organized and ready to apply machine learning algorithms. Then, in the second part, the reinforcement learning method inspired by voting systems starts working. Finally, in the third part of the framework, learning is done to choose the classification algorithm, and prediction is also presented.
The structure of this section includes three subsections. In subsection A, the first step of the Fourth Degree Learning is explained. In subsection B, the FDL method, which is the main topic of this article, is fully explained. Finally, sub-section C explains the last step of the framework presented.

### *A. Data Cleaning*

In the first stage of this framework, four main tasks are performed on the dataset. These steps include Privacy Detection, Null Detection, Duplicate Detection, and Outlier Detection. A full explanation of each is provided below.

- ***Privacy Detection***

Since the general vision for the production of this framework was to work on different types of datasets, in this section, if the dataset has personal characteristics such as national number, address, phone number, and such items, it will be removed. Features such as patient IDs that are useless will also be removed. Also, data figures are extracted so that the features can be intuitively analyzed.

- ***Duplicate & Outlier Detection***

In order to have data that is ready for use in all respects, at this stage, duplicate samples have been extracted first. After extracting them, the features that have outliers have also been determined. Then, these two problems are combined, and duplicative and high-quality examples are removed.

- ***Null Detection***

The possibility of invalid data in real datasets is high. That is why all invalid data are extracted in this section. After that, the available null data is deleted, and the data balance is checked. If the data set and its target are not balanced, the data is moderated with balancing methods and moved to the next stage. The classifiers we use are, Decision Tree [18] , Random Forest [19] and Support Vector Classifier [20] and the shark smell optimization algorithm is used for this balancing [31].

### B. Fourth Degree Learning (FDL)

The main part of this research is presented in this section. To solve the time-consuming problem of which classification algorithm is suitable for a data set, this method has been implemented. The Fourth Degree Learning method learns from repeating a loop and scoring. Then, the work of the framework begins with a voting algorithm which is roulette wheel.

In the first step of the iteration loop, the sorted data sets are fed to the classifier algorithms. The best algorithm is selected using the roulette wheel. Finally, the evaluation and scoring system is implemented. The first step is to run the algorithms.

The algorithms implemented in this section are: Decision Tree Algorithm [18], Random Forest Algorithm [19] and Support Vector Classifier Algorithm [20].

In this algorithm, there is a list of classifiers called $AlgL$. The components of this list are displayed as $AlgL = \{Alg_1, Alg_2, Alg_3, \ldots, Alg_n\}$ . In this algorithm, all classifiers must be run once. This initial implementation is done because initial scoring is applied to them. The policy used in this reinforcement learning algorithm is to achieve the highest possible score. The scoring of each algorithm is determined according to equation 1. In this regard, the score is equal to the Fourth Degree Learning rate. To understand SL, precision (PR), recall (R), F-score (Fs) and accuracy (Acc) metrics have been used.

$$SL = \frac{Pr + R + Fs + Acc}{100} \quad (1)$$

FDL is calculated for each algorithm and stored in order from the maximum mean of this parameter to the minimum in a list called the Algorithm Reward ($AlgR$). The noteworthy point is that all algorithms must have a score in each iteration. For this reason, the parameters of the current iteration and overall iteration are introduced with Ci and i. In addition, the scoring condition for each algorithm is given in equation 2. In this equation 2, the condition is that the score belongs to Ward's algorithm if it has the Maximum value of FDL and is in the current iteration (Ci). As the iteration steps in this algorithm continue, the scoring should be updated. Algorithm Rewards ($AlgR$) are stored in a list called Reward List ($RList$) and are updated at the end of each iteration. The best algorithm with the highest score is indicated by the Max Fourth Degree Learning rate ($Alg^{SL}_{Max}$) parameter. In fact, the total scores of each algorithm are stored in the list of variables, which are added in equations 3 and 4.

$$AlgR = \{\, AlgR + 1 \mid Alg_{SL} = Maximum, Iteration = Ci \,\} \quad (2)$$

$$AlgR = \sum_{i=1}^{n} (Alg^{SL}_{Max} + 1) \quad (3)$$

$$RList = \{\, List\ Alg \mid Maximum\ SL > Alg > Minimum\ SL \,\} \quad (4)$$

In order to choose a new algorithm, the algorithms are listed from the best to the worst using the $RList$. In this algorithm, in order to avoid the greedy selection of the best algorithm using a roulette wheel, all algorithms have a chance to be selected until the last stage of repetition. Roulette Wheel output is introduced with the $RW$ parameter, which stands for Roulette Wheel. Equation 5 shows the roulette wheel. In this formula, the best algorithm is divided by the total scores of other algorithms. This process also gives chance to weak algorithms. The final stage is the final learning, which will be selected and implemented according to the mentioned process of the algorithm output from the roulette wheel, whose order is given in Equation 6. Ai represents the Action for each Iteration and is equal to the execution of the selected algorithm according to the roulette cycle. This process happens until the condition stops it. In this framework, the condition for stopping is the number of iterations.

$$RW = \frac{Alg^{SL}_{Max}}{\sum_{Alg=1}^{n} Alg_{SL}} \quad (5)$$

$$Ai = \left\{ Run\ Selected\ Alg \,\middle|\, \frac{Alg^{SL}_{Max}}{\sum_{Alg=1}^{n} Alg_{SL}} \right\} \quad (6)$$

### C. Steps of Fourth Degree Learning (FDL)

In this section, the main steps of the framework are presented. This framework consists of 3 main parts: Data Cleaning, Reinforcement Learning, and presentation of results and predictions. The steps are represented:

- Reading data.
- Prepare data (Privacy Detection, Null Detection, Duplicate Detection, and Outlier Detection).
- Implementation of algorithms and list adjustment.
- Importing data to all algorithms and execution.
- Choosing the best rhythm pattern with the roulette cycle.
- Scoring the selected algorithm.
- Checking the stop condition (if it is not met, it will be repeated from step 3).
- Providing the best algorithm and prediction.

## 4. EXPERIMENTAL EVALUATION

This section reports our experiments to evaluate the proposed framework. This part supplies detailed information on Balancer Selection.

The results indicate a better performance for the both majority and minority classes, which is the Class 1 (Stage 1,2 and 3) and class 2 (Stage 4) of PBC. However, the focus of this work is to provide an accurate classification for both classes. In particular, we use Decision Trees (DT) [18], Random Forest (RF) [19], and Support Vector Machine (SVM) [20] classifiers.. All initial parameters for each classifier are set to the default values.

Table I indicates the evaluation metrics, F1-score, precision, and recall, for SSO balancer and three classification methods. The rest of the table indicates the classification performance when the SSO balancer methods balance the data. SSO affect the evaluation results, especially the minority class, the Stage 4 group. As the results show, classification over the data balanced by SSO results in a better performance in contrast to the Raw Data. RF has the highest F1-score in comparisons with the other classification methods. For the other evaluation metrics, the results of this classifier are approximately the best. It is worth mentioning that some classifiers have a higher evaluation metric than RF. However, the overall performance of this method for both classes of class1and stage 4 is higher than the others.

TABLE I. F1-SCORE (FS), PRECISION (PR), RECALL (R), FOR THREE CLASSIEFIRES OVER IMBALANCED DATA

| Balancer | Classifier | PR | R | Fs | Acc | SL |
|---|---|---|---|---|---|---|
| Raw Data | DT | 85% | 15% | 75% | 62% | 2.37 |
| | SVC | 83% | 49% | 81% | 53% | 2.66 |
| | RF | 86% | 52% | 82% | 63% | 2.83 |
| **SSO** | DT | 95% | 94% | 95% | 94% | 3.78 |
| | SVC | 95% | 94% | 95% | 94% | 3.78 |
| | **RF** | **98%** | **96%** | **97%** | **96%** | **3.87** |

Proving that the Right Choice of Classification Algorithms Based on Fourth Degree Learning is in the right way is discussed. In other words, this proof is about the advantage of the SL. Table II provides, information about the evaluation results for each iteration in the FDL and all classifier's accuracy with a balanced dataset. The FDL is based on high to low, which means the highest iteration (Iter) is the best. The best accuracy in Iteration 80 and above it, is in the range of 95%.

TABLE II. EVALUATION RESULTS OF RIGHT CHOICE OF CLASSIFICATION ALGORITHM BASED ON FOURTH DEGREE LEARNING (SL)

| Iter | Iter 80 | Iter 70 | Iter 60 | Iter 50 | Iter 40 | Iter 30 | Iter 20 | Iter 10 |
|---|---|---|---|---|---|---|---|---|
| SVC | 95% | 81% | 77% | 77% | 75% | 81% | 75% | 76% |
| **RF** | **98%** | **95%** | **89%** | **88%** | **85%** | **85%** | **76%** | **71%** |
| DT | 93% | 77% | 72% | 73% | 74% | 79% | 73% | 73% |

# 5. CONCLUSION

This article describes the early prevention of PBC based on Stages 1 to 4. We propose a novel reinforcement learning method called Fourth Degree Learning to select the best classifier for the data. We provide several experiments to evaluate the framework using both balanced and imbalanced data. The results indicate that our framework predicts Stage 4 with an accuracy of 98%. The novelties of this work could provide inspiration for researchers in both machine learning and medical studies. For future work, we plan to evaluate our framework using different big data and study its performance for other types of diseases in the healthcare field. Improving the set of balancers and classifiers in the framework could be also another direction for future work. An online version of the proposed framework could be implemented and used for runtime PBC prediction.